%% file: main.tex
\documentclass[runningheads]{llncs}

\usepackage[table,dvipsnames]{xcolor}
\usepackage{eccv}
\usepackage{eccvabbrv}

\authorrunning{L. Feng et al.}

\usepackage{graphicx}
\usepackage{amsmath}
\usepackage{amssymb}
\usepackage{booktabs}
\usepackage{enumitem}
\usepackage{adjustbox} 
\usepackage{multirow}

\usepackage[accsupp]{axessibility}

\usepackage{pgfplots}
\pgfplotsset{compat=1.17}

\usepackage[pagebackref,breaklinks,colorlinks]{hyperref}

\definecolor{Blue}{HTML}{4169E1}
\definecolor{ForestGreen}{HTML}{CD5C5C}
\definecolor{GreenYellow}{HTML}{3CB371}

\usepackage[capitalize]{cleveref}
\crefname{section}{Sec.}{Secs.}
\Crefname{section}{Section}{Sections}
\Crefname{table}{Table}{Tables}
\crefname{table}{Tab.}{Tabs.}

\usepackage{colortbl} 

\newcommand{\waymo}{WOMD}

\begin{document}

\title{UniTraj: A Unified Framework for \\ Scalable Vehicle Trajectory Prediction}

\author{Lan Feng\inst{*,1} \and
Mohammadhossein Bahari\inst{*,1} \and
Kaouther Messaoud Ben Amor \inst{1} \and Éloi Zablocki \inst{2} \and Matthieu Cord \inst{2,3} \and Alexandre Alahi \inst{1}}


\institute{EPFL, Switzerland, \\
\email{firstname.lastname@epfl.ch}\and
Valeo.ai, France, \\ 
 \and Sorbonne Université, France }
 
\maketitle
\let\thefootnote\relax\footnotetext{\leftline{$^*$ Equal contribution as the first authors. }}

\begin{abstract}
Vehicle trajectory prediction has increasingly relied on data-driven solutions, but their ability to scale to different data domains and the impact of larger dataset sizes on their generalization remain under-explored. 
While these questions can be studied by employing multiple datasets, it is challenging due to several discrepancies, \textit{e.g.,} in data formats, map resolution, and semantic annotation types. To address these challenges, we introduce UniTraj, a comprehensive framework that unifies various datasets, models, and evaluation criteria, presenting new opportunities for the vehicle trajectory prediction field.
In particular, using UniTraj, we conduct extensive experiments and find that model performance significantly drops when transferred to other datasets.
However, enlarging data size and diversity can substantially improve performance, leading to a new state-of-the-art result for the nuScenes dataset. We provide insights into dataset characteristics to explain these findings.
The code can be found here: \hyperlink{https://github.com/vita-epfl/UniTraj}{https://github.com/vita-epfl/UniTraj}.

\end{abstract}

\begin{figure}[t]
    \centering
    \includegraphics[width=0.85\textwidth]{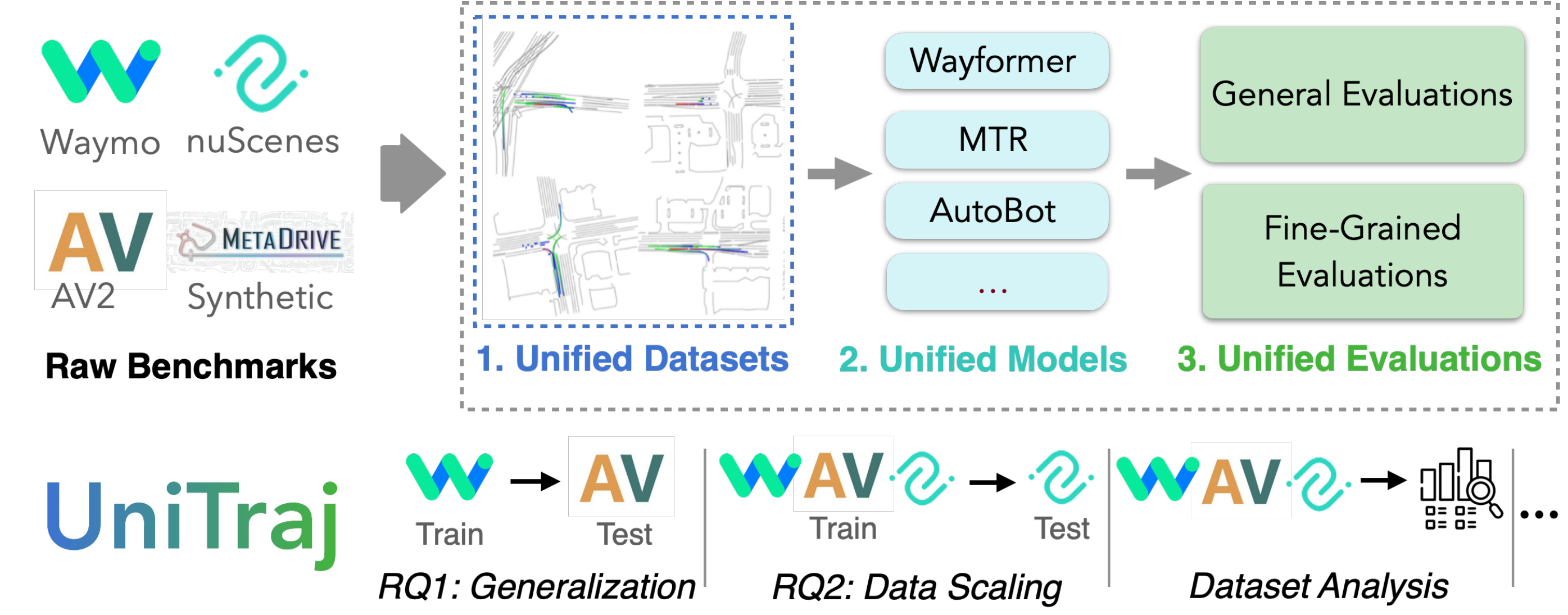}
    \caption{\textbf{UniTraj framework.} The framework unifies various datasets, forming the largest vehicle trajectory prediction dataset available. It also includes multiple state-of-the-art prediction models and various evaluation strategies, making it suitable for trajectory prediction experimentation.
    The framework enables the study of diverse Research Questions, including (RQ1) the generalization of trajectory prediction models across different domains and (RQ2) the impact of data size on prediction performance.
    }
    \label{fig:teaser}
\end{figure}

\section{Introduction}
\label{sec:intro}

Predicting the trajectories of surrounding vehicles is essential for ensuring the safety and collision avoidance of autonomous driving systems. With the advent of deep learning, researchers have turned to data-driven solutions to tackle this prediction task. However, while these models can achieve high accuracy, they are heavily reliant on the specific data domain used for training.

An autonomous driving system may encounter various situations such as diverse geographical locations. These various situations introduce data domain shifts, which can significantly impact the performance of the prediction models. Consequently, it is essential to study the performance of the models across diverse domains, such as datasets and cities. However, despite the importance of the question, the generalization of models to different domains has not been adequately studied yet. Therefore, our first Research Question (RQ1) is to investigate the performance drop of trajectory prediction models when transferred to new domains. 

A potential solution to improve the generalization ability of prediction models is to scale up the sizes of the datasets to cover a broader spectrum of driving scenarios. While there is a trend in extending datasets' sizes~\cite{chang2019argoverse,ettinger2021waymo,caesar2021nuplan,wilson2023argoverse2}, the impact of dataset size on the performance of trajectory prediction models remains largely unexplored. Our second research question (RQ2) is then to study the impact of increasing dataset sizes on the performance of the prediction models.

Exploring these two research questions involves leveraging multiple trajectory prediction datasets.
Firstly, these datasets provide diverse domains, allowing for a thorough examination of model generalization across different domains (RQ1).
Secondly, combining these datasets creates a much larger dataset, enabling an exploration of the asymptotic limits of data scaling (RQ2).
However, significant challenges exist when attempting to leverage multiple datasets.
(1) Each of these datasets has a unique data format, posing practical difficulties for researchers utilizing multiple datasets. (2) Each of the datasets undergoes collection and annotation through distinct strategies, with semi-automatic pre-annotations and manual curations \cite{caesar2020nuscenes,ettinger2021waymo,caesar2021nuplan}. This leads to multiple discrepancies such as variations in resolution, sampling rates, and types of semantic annotations.
(3) Comparing model performance across datasets is not straightforward due to varying dataset settings (e.g., prediction horizons) and evaluation metrics (e.g., mAP metric is used in \waymo{} \cite{ettinger2021waymo} and brier-FDE metric in Argoverse~2 \cite{wilson2023argoverse2}). In short, while each of the datasets contributes to the progress in the field, they have been developed independently, without considering harmonization with existing ones. As a result, many trajectory prediction studies train and evaluate their models using a single dataset~\cite{coscia2018long,liang2020learning,bahari2021injecting,cao2022advdo,shi2022motion_mtr,bahari2022vehicle,pourkeshavarz2023learn,hsu2023responsibility,jiang2023motiondiffuser,bhattacharyya2023ssl, Messaoud_00}.

To tackle these challenges, we introduce `UniTraj', a comprehensive vehicle trajectory prediction framework.
UniTraj seamlessly integrates and unifies multiple data sources (including nuScenes~\cite{caesar2020nuscenes}, Argoverse~2~\cite{wilson2023argoverse2} and Waymo Open Motion Dataset - \waymo{} ~\cite{ettinger2021waymo}), 
models (including AutoBot~\cite{girgis2022autobot}, MTR~\cite{shi2022motion_mtr}, and Wayformer~\cite{nayakanti2023wayformer}), and evaluations.
UniTraj not only serves as a solution to tackle our research questions but also provides a comprehensive and flexible platform for the community. 
First, it is designed for the effortless inclusion of new datasets by proposing a unified data structure compatible with various datasets.
Second, Unitraj supports and simplifies the integration of new methods by providing numerous essential data processing and loss functions relevant to the trajectory prediction task.
Lastly, UniTraj offers unified evaluation metrics, as well as diverse and insightful evaluation approaches, such as analyzing performance on the long-tail data instances as well as different clusters of data samples to allow a more in-depth understanding of model behavior. \Cref{fig:teaser} shows the overview of the framework.

We conduct extensive experiments using the UniTraj framework to shed light on our two research questions.
Our findings reveal a large performance drop when transitioning between data sources, alongside variations in the generalization abilities induced by different datasets (RQ1).
We also show that scaling up dataset size and diversity can enhance model performance significantly without any architectural modifications, leading us to rank $1$\textsuperscript{st} in the nuScenes public leaderboard. 
This is accomplished by training models on all existing datasets in the framework. This unified dataset forms the largest public data one can use to train a vehicle trajectory prediction model, with more than $2$M samples, $1337$ hours of data, and $15$ different cities.
Finally, by providing an in-depth analysis of the datasets, we offer a more comprehensive understanding of their characteristics. Our analysis reveals that the datasets' generalization capabilities are not only attributed to their size, but also their intrinsic diversity.
We believe that the framework opens up new opportunities in the trajectory prediction field, and we will release the framework to foster further advancements. In summary, our contributions are as follows:
\begin{itemize}
    \item We introduce UniTraj, a comprehensive open-source framework for vehicle trajectory prediction, integrating various datasets, models, and evaluations. It offers a unified platform for comprehensive research in this field.
    \item We investigate models' generalization across different datasets and cities and provide insight into the characteristics of datasets on which models acquire better generalization capacities.
    \item We explore the data scaling impact on model performance employing the largest collection of datasets currently available, and establish a new state-of-the-art model on the nuScenes dataset.
    \item Finally, we provide an in-depth comparative analysis of the datasets, shedding light on our experimental findings.
\end{itemize}

\section{Previous work}

\noindent\textbf{Trajectory prediction datasets.}
Many academic and industrial laboratories have paved the way for research development by open-sourcing real-world driving datasets~\cite{robicquet2016learning,zhan2019interaction,bock2020ind,ettinger2021waymo,caesar2020nuscenes,chang2019argoverse,houston2021lyft,malinin2021shifts,caesar2021nuplan,wilson2023argoverse2}. 
Notably, Argoverse \cite{chang2019argoverse} was among the pioneers in releasing the lane graph information, nuScenes \cite{caesar2020nuscenes} expanded the variety of scenes, Waymo \cite{ettinger2021waymo} enriched their dataset with fine-grained information, and 
recently, Argoverse~2 \cite{wilson2023argoverse2} released the largest data in terms of unique roadways.
While these datasets contribute to field developments, they have been developed in isolation without considering harmonization with former datasets. Thus, there exist multiple challenges in combining them due to various incompatibilities.
This work addresses the challenges through a unified framework.

\noindent\textbf{Trajectory prediction benchmarks.} 
Multi-dataset benchmarks have already been explored in various domains such as object detection \cite{zhou2022simple}, semantic segmentation \cite{kim2022learning,shi2021multi}, and pose prediction \cite{saadatnejad_pose}. 
In the field of trajectory prediction, such benchmarks have primarily been developed for human trajectory prediction \cite{sadeghian2018trajnet,amirian2020opentraj,rudenko2022atlas}. Notably, Trajnet++ \cite{kothari2021human} provides an interaction-centric benchmark by categorizing trajectories based on the presence of an interaction. trajdata \cite{ivanovic2023trajdata} is a unified interface to multiple human trajectory datasets incorporating scene context into the inputs.  
A related work for the task of vehicle trajectory planning is ScenarioNet \cite{li2023scenarionet}, a simulator aggregating multiple real-world datasets into a unified format and providing a planning development and evaluation framework. 
To the best of our knowledge, we are the first to propose an open-source framework for vehicle trajectory prediction. 
Our framework is not limited to including multiple datasets; it also integrates a variety of trajectory prediction models and evaluation methodologies, thereby providing a comprehensive resource for advancing research and development in the vehicle trajectory prediction task.

\noindent\textbf{Generalization of trajectory prediction models.}
The discrepancies in data formats in vehicle trajectory prediction datasets hinder research on cross-dataset generalization, leading to limited studies in this area.
In \cite{ye2023improving},  one dataset is divided into different domains to explore model generalization. 
Authors in \cite{shao2023does} propose an epistemic uncertainty estimation approach and perform cross-dataset evaluation. In \cite{gilles2022uncertainty}, the authors studied cross-dataset generalization of models and showed a performance gap between datasets. However, they provide limited insights into the sources of the generalization gap. Moreover, their code is not publicly available. 
Previous works also investigated some generalization aspects of trajectory prediction models when they deal with new scenes and cities \cite{chen2021counterfactual,kothari2022mosa,liu2022causal}, 
new agent types \cite{kothari2022mosa}, using perception outputs instead of curated annotations \cite{weng2022affinipred,zhang2022forecast_from_detection,xu2024end2endforecast}, and facing adversarial situations \cite{bahari2022vehicle,sarva2023adv3d,saadatnejad2021sattack}.
In this work, we conduct more extensive and in-depth cross-dataset, and cross-city analyses as well as multi-dataset training. Moreover, we provide insights into the dataset characteristics, explaining the findings. We also release an open-source framework to facilitate this line of research.

\section{UniTraj framework}
The UniTraj framework, illustrated in \Cref{fig:teaser}, consists of three main components.
The first component unifies the format and features of various datasets (see \Cref{sec:unified_data}).
The second component adapts trajectory prediction models to the unified data format, facilitating their training (see \Cref{sec:unified_models}).
The final component consists of a comprehensive and shared evaluation process for the models (see \Cref{sec:unified_evaluation}).
The integration of the components allows for diverse experimentation, such as cross-dataset training, evaluation, and dataset analysis.

\begin{table*}[t]
\centering
\caption{\textbf{Summary of the discrepancies in data features.} The table shows the features for each dataset and the unified version of the features.  Most of the unified features are flexible and can be chosen by the user.}
\resizebox{\textwidth}{!}{
\begin{tabular}{l l *{4}{@{\hspace{.3cm}}c}}
\toprule
     & & \multicolumn{1}{c}{Argoverse2}  & \multicolumn{1}{c}{\waymo{}} & \multicolumn{1}{c}{nuScenes}  & \multicolumn{1}{c}{\textbf{UniTraj}}                     \\
\midrule
\textbf{Coordinate frame} & & Scene-centric & Scene-centric & Scene-centric & Agent-centric         \\ 
\multirow{2}{*}{\textbf{Time length}} & Past & 5 sec & 1 sec & 2 sec & [0 - 8] sec \\
& Future & 6 sec & 8 sec & 6 sec & [1 -  8] sec \\[.3cm] 
\multirow{3}{*}{\textbf{Agent features}}  & Annotations & velocity, heading & velocity, heading & velocity, heading & velocity, heading \\
& Other info & --- & bounding box size & --- & acceleration \\
& Coordinates & 2D & 3D & 2D & 2D \\[.3cm]
\multirow{3}{*}{\textbf{Map features}}  & Range & $\sim$200m & $\sim$200m & $\sim$500m & [0 - 500] m\\
& Resolution & 0.2m$\sim$2m & $\sim$0.5m & $\sim$1m & [0.2 - 2] m\\
& Coordinates & 2D & 3D & 2D & 2D \\[.2cm]

\bottomrule
\end{tabular}}

\label{tab:data_features}
\end{table*}

\subsection{Unified data}
\label{sec:unified_data}

Two types of discrepancies are found across trajectory forecasting datasets: \emph{data formats} and \emph{data features}.
The former amounts to differences in the way data is structured and organized, while the latter stems from differences in the characteristics of the data itself, such as spatio-temporal resolution, range, and agent and map annotation taxonomy.
In this section, we present solutions to tackle both types of discrepancies.

\subsubsection{Unified data format:}
To address the issue of different data formats used in trajectory prediction datasets, such as TFRecord in \waymo{}~\cite{ettinger2021waymo} and Apache Parquet in Argoverse~2~\cite{wilson2023argoverse2}, we utilize ScenarioNet~\cite{li2023scenarionet}. 
ScenarioNet was initially designed for traffic scenario simulation and modeling, but we repurposed it for the vehicle trajectory prediction task.
It provides a unified scenario description format containing HD maps and detailed object annotations, which simplifies the process of decoding the dataset with different formats.
ScenarioNet currently supports converting \waymo{}, nuScenes, and nuPlan, and we extend its support to Argoverse~2 for our research. 
This reduces the need for multiple versions of preprocessing code to extract information from raw datasets and create batched data for the training of prediction models.

\subsubsection{Unified data features:}
\label{sec:unified_features}
Despite the data being converted into a unified format, significant discrepancies persist across the datasets, affecting model performance.  
For example, the scenarios are $11$ seconds long in Argoverse~2, while they are $9$ seconds in \waymo{}; or the precision of map annotations are $1$ meters in nuScenes while they are $0.5$ meters in \waymo{}.
Therefore, we aim to harmonize these discrepancies and minimize their impact on the model's performance.
\Cref{tab:data_features} summarizes the discrepancies and the unified features.
Our data processing approach involves specific harmonizations, including the following:
\\
\begin{itemize}
    \item \textbf{Coordinate Frame.}
    Recent trajectory prediction models predominantly utilize vectorized, agent-centric data as input~\cite{shi2022motion_mtr, girgis2022autobot, nayakanti2023wayformer, varadarajan2022multipath++, zhou2023query}. Our data processing pipeline is designed to transform scene-level raw data into this format. It processes traffic scenarios, which consist of multiple trajectories, and selects trajectories designated as training samples within the datasets. The pipeline then converts the entire scenario into the coordinate frames of these selected agents, and normalizes the input accordingly.

\item \textbf{Time Length.} The trajectories in different datasets are with the same frequency of $10$Hz but with a duration ranging from 8 to 20 seconds. 
To standardize this aspect, we truncate all trajectories to a uniform length of 8 seconds. Within this duration, UniTraj provides the option to flexibly determine a unified length of past and future trajectories for all datasets. 

\item \textbf{Agent Features.} Among the datasets, \waymo{} 
 provides the most detailed agent information, including 3D coordinates, velocity, heading, and bounding box size, whereas nuScenes lacks certain data, like bounding box size. We standardize inputs across datasets by using 2D coordinates, velocity, and heading. Our data processing also introduces supplementary features, such as one-hot encoding of agent type and time steps of trajectories, and acceleration. These elements are combined to create a rich, unified input for the model.

\item \textbf{Map Features.} 
Datasets differ in HD map resolution. We normalize this by using linear interpolation to standardize the distance between consecutive points to 0.5 meters, with an option for further downsampling to adjust map resolution. Additionally, we enrich the data with each lane point's direction and one-hot encode lane types. Our experiments utilize semantic map classes such as center lanes, road lines, crosswalks, speed bumps, and stop signs.
\end{itemize}

Our framework allows for customization of specific features through predefined parameters for focused single-dataset research, while still providing a standardized data format across all datasets. The data processing module supports various parameters, such as the length of historical and future trajectories, number of points per lane, map resolution, types of surrounding agents and lines, and masked attributes.
Thanks to its modular structure, our data processing pipeline enables easy integration of new processing methodologies and models. The framework is equipped with multi-processing and caching mechanisms for efficient processing.
Our framework currently includes four large-scale, real-world datasets with over 1337 hours of driving data from 15 cities.

\subsection{Unified models}
\label{sec:unified_models}

Trajectory prediction models are often implemented in different pipelines, making direct comparisons challenging and fairness hard to ensure. We integrate three recent trajectory prediction models within the UniTraj framework.
These models were chosen based on their state-of-the-art results on various benchmarks, indicating the research value of their designs.
We include: 
\begin{itemize}
    \item \textbf{AutoBot} \cite{girgis2022autobot} is a recent transformer-based model with competitive results on multiple existing datasets. It is based on equivariant feature learning to learn the joint distribution of trajectories with multi-head attention blocks. 

    \item \textbf{MTR} \cite{shi2022motion_mtr} ranked first on the \waymo{} Challenge in 2022. It operates by integrating global intention priors with local movement refinement. It uses a limited number of adaptable motion query pairs, allowing precise trajectory prediction and improvement for different motion types.
    
    \item \textbf{Wayformer} \cite{nayakanti2023wayformer} is a transformer-based model, featuring a multi-axis encoder.
    It employs latent queries that facilitate the combination of multi-dimensional inputs. We re-implement the model, as the original code has not been released.
\end{itemize}

The models' capacities cover a large range ($1.5$M parameters for AutoBot, $60.1$M for MTR, and $16.5$M for Wayformer), enabling research on model size and scaling. \\
\noindent \textbf{Integrating new models:}
The flexibility of UniTraj's data processing pipeline greatly simplifies the integration of new models.
Furthermore, we provide a standardized output format, enabling seamless use of UniTraj's evaluation and logging tools.

\subsection{Unified evaluation}
\label{sec:unified_evaluation}
In trajectory prediction, various metrics have been proposed to evaluate the models, yet there is no consensus about them.
As a result, each dataset provides a different set of evaluation metrics, making it challenging to compare performances across datasets.
For example, \waymo{} employs mean average precision (mAP) metric \cite{ettinger2021waymo} while Argoverse~2 uses brier minimum Final Displacement Error (brier-minFDE) \cite{wilson2023argoverse2}.
Our framework provides a unified set of metrics to allow comprehensive and consistent evaluation across different datasets.
To this end, we employ two sets of metrics: general and fine-grained evaluation metrics.

\subsubsection{General evaluations:}
The most common metrics in the literature are the ones that provide an overall score based on accuracy measures by comparing the output with the ground truth in different aspects. 
We include the following three general metrics in the framework:
\\
1)~{Minimum Average / Final Displacement Error (minADE/minFDE):} It represents the minimum average/final displacement error between the predictions and the ground truth. The minimum is computed over the $6$ modes of the output.\\ 
2)~{Miss Rate (MR):} It is defined as the ratio of the samples with minFDE exceeding $2$ meters, and is useful where up to $2$ meters deviation is acceptable.\\
3)~{Brier Minimum Final Displacement Error (brier-minFDE):} While the previous metrics focus on covering the ground truth, they do not account for the probability assigned to each predicted trajectory. The brier-minFDE metric addresses this by adding a penalty term, $(1 - p)^2$, to the minFDE where $p$ corresponds to the probability of the trajectory that best matches the ground truth.

\noindent\textbf{Fine-grained evaluations:}
We also provide two fine-grained evaluations.
\\
\noindent\textbf{(1) Trajectory types.}
Datasets usually exhibit a significant prevalence of `straight' trajectories, resulting in heavily imbalanced datasets. Besides, we argue that rare trajectory types can sometimes be the more safety-critical ones.
Therefore, it is critical to specifically access prediction performances on rare situations and trajectory types.
To address this, the UniTraj framework enables the stratification of evaluation metrics based on trajectory types. In practice, we adopt the trajectory taxonomy defined in the \waymo{} challenge \cite{ettinger2021waymo}, to categorize trajectories into the following groups: `stationary', `straight', `straight left', `straight right', `left-turn', `right-turn', `left u-turn', `right u-turn'.
While the use of this taxonomy provides valuable insights, its scope has limitations as it does not account for variations in motion dynamics.
For instance, it does not differentiate between straight accelerating and decelerating trajectories, both of which are categorized as `straight'.
Consequently, we additionally use the notion of `Kalman difficulty' introduced below.

\noindent\textbf{(2) Kalman difficulty.}
Some situations are more challenging to forecast than others, typically when the future is not a simple extrapolation of the past and when contextual factors play a significant role. The context encloses various elements such as map data, social interactions, or input signals coming from perception.
Moreover, previous works \cite{makansi2021challenging_long_tail,benyounes2022raising_context,wang2023fend} observe that these complex scenarios, while critical, are much less frequent than scenarios that are easier to forecast.
To specifically evaluate the performance over critical cases, and reduce evaluation noise coming from the large number of simple scenarios, the authors in~\cite{makansi2021challenging_long_tail} propose to filter them as the ones with a high mismatch between their ground truth and predictions from a Kalman filter \cite{kalman1960new}.
We follow this idea as it offers a simple method to evaluate how challenging a situation is.
Accordingly, UniTraj stratifies evaluation metrics based on \emph{Kalman difficulty} that we define as the FDE between the ground-truth trajectory and the prediction of a linear Kalman filter.

\section{Experiments}
The UniTraj framework opens up new opportunities for research and experimentation. This section presents experiments highlighting these opportunities, focusing on cross-domain (i.e., cross-dataset and cross-city) generalization (RQ1) in \Cref{sec:generalization_evaluation}, and data scaling impact (RQ2) for trajectory prediction models in \Cref{sec:multi_dataset_training}. 
We provide fine-grained dataset analyses and discussions in \Cref{sec:interpreting}. Additional experiments in the appendix, such as continual learning and synthetic-to-real transfer, further demonstrate the framework's research utility.

\noindent\textbf{Experimental settings:}
\label{sec:experiment_settings}
We replicate the model configurations and hyper-parameters from their original implementations. 
Throughout the experiments, we have limited the training and validation samples to vehicle trajectories. The map range extends to a $100$m radius with a spatial resolution of $0.5$m. The temporal parameters are set to $2$ seconds of historical trajectories and $6$ second future trajectories.
For our multi-dataset training experiments, we utilize \waymo{}~\cite{ettinger2021waymo}, Argoverse~2~\cite{wilson2023argoverse2}, and nuScenes~\cite{caesar2020nuscenes} datasets.
Since the nuPlan~\cite{caesar2021nuplan} dataset is oriented towards planning tasks and lacks an official training/validation set for prediction tasks, we exclusively use it for the cross-city generalization studies due to its large number of samples for different cities.
We only report the results with the brier-minFDE metric and leave other metrics for the appendix.

\subsection{Generalization evaluation}
\label{sec:generalization_evaluation}

Generalization to new domains is a crucial challenge for data-driven models, necessitating diverse data for comprehensive evaluation. The UniTraj framework enables the exploration of model generalization across various datasets and cities.

\subsubsection{Cross-dataset evaluation:}
\label{sec:cross_dataset_evaluation}
To assess the generalization capabilities of models, we train models on each individual dataset and evaluate their performance on all other available datasets.
The findings are presented in \Cref{tab:cross_dataset_evaluation}. 
Analyzing the data in different columns of the table, the first observation is that all models' performances decline significantly when models are tested on other datasets. This is a consistent trend across all of the three model architectures, and all of the considered datasets. 
For instance, the second column under MTR reports the performance evaluated on the validation set of Argoverse~2. It indicates that MTR achieves its peak performance when it is trained on the training set of Argoverse~2 itself, while models trained on nuScenes and \waymo{} exhibit significantly lower performances.

With a more detailed investigation, we can also compare the generalization capabilities of different datasets.
For instance, considering the same column, the model trained on \waymo{} outperforms the one trained on nuScenes when evaluated on the Argoverse2 dataset.
By making similar comparisons across other columns, we establish a generalization order: models trained on \waymo{} data exhibit the highest generalization ability, followed by those trained on Argoverse2, and then nuScenes.
This order remains consistent across all models.
The superior generalization of models trained on \waymo{} can be attributed to both the larger number of data samples and the greater variety present in the \waymo{} dataset. We provide a more detailed explanation in Section~\ref{sec:interpreting}, where we discuss the specific characteristics of each dataset and their influence on model performance and generalization.

\begin{table*}
\caption{\textbf{Cross-dataset generalization and multi-dataset training experiments.} Training and validation are across multiple datasets. Rows indicate the \colorbox{blue!15}{training data} of the model, columns indicate the \colorbox{orange!25}{evaluation data}. `All' designates the combination of the three considered datasets. 
The study is conducted for three models (AutoBot \cite{girgis2022autobot}, MTR \cite{shi2022motion_mtr}, and Wayformer \cite{nayakanti2023wayformer}). `$\ast$' indicates our internal implementation of the model. We report the brier-minFDE ($\downarrow$) metric.
}
\resizebox{1.0\textwidth}{!}{
\begin{tabular}{@{}l r c *{8}{@{\hspace{0.05cm}}c} @{}}
\toprule
& & \multicolumn{3}{c}{\text{MTR \cite{shi2022motion_mtr}}} & \multicolumn{3}{c}{\text{Wayformer * \cite{nayakanti2023wayformer}}} & \multicolumn{3}{c}{\text{AutoBot \cite{girgis2022autobot}}} \\
\cmidrule(lr){3-5} \cmidrule(lr){6-8} \cmidrule(lr){9-11}
& & \multicolumn{9}{c}{\cellcolor{orange!25} $\leftarrow$ \textbf{Evaluation} $\rightarrow$}\\
\cellcolor{blue!15}{$\downarrow$ \textbf{Training}} & \#\text{trajs} & \cellcolor{orange!25} \text{nuScenes}  & \cellcolor{orange!25} \text{Argoverse~2} & \cellcolor{orange!25} \text{\waymo{}} & \cellcolor{orange!25} \text{nuScenes}  & \cellcolor{orange!25} \text{Argoverse~2} & \cellcolor{orange!25} \text{\waymo{}} & \cellcolor{orange!25} \text{nuScenes}  & \cellcolor{orange!25} \text{Argoverse~2} & \cellcolor{orange!25} \text{\waymo{}}\hspace{-0.3cm}\\
\midrule
\cellcolor{blue!15} \text{nuScenes} & 32\text{k} & \textbf{2.86} & 4.50 & \multicolumn{1}{c|}{7.38} & \textbf{3.06} & 4.68 & \multicolumn{1}{c|}{7.16} & \textbf{3.36} & 4.48 & 6.89 \\
\cellcolor{blue!15} \text{Argoverse~2} & 180\text{k} & 3.72 & \textbf{2.08} & \multicolumn{1}{c|}{4.68} & 3.69 & \textbf{2.38} & \multicolumn{1}{c|}{4.80} & 4.35 & \textbf{2.51} & 4.43 \\
\cellcolor{blue!15} \text{\waymo{}  } & 1800\text{k} & 3.10 & 3.63 & \multicolumn{1}{c|}{\textbf{2.13}} & 3.12 & 3.60 & \multicolumn{1}{c|}{\textbf{2.10}}  & 3.73 & 3.23 & \textbf{2.47} \\
\midrule
\cellcolor{blue!15} \text{All} &  2012\text{k} & \textbf{2.27} & \textbf{1.99} & \multicolumn{1}{c|}{\textbf{2.13}} & \textbf{2.32} & \textbf{2.12} & \multicolumn{1}{c|}{\textbf{2.09}}  & \textbf{3.07} & 2.54 & \textbf{2.47} \\
\bottomrule
\end{tabular}}
\label{tab:cross_dataset_evaluation}
\end{table*}
\subsubsection{Cross-city evaluation:}
\label{sec:cross_cities_evaluation}
Despite our care to standardize data formats and align features among datasets, certain fundamental discrepancies may persist caused by the data collection and annotation processes. For instance, annotation noises could still exist across datasets.
To control for this potential residual discrepancy, we explore the generalization of AutoBot when the city is changed inside a single dataset.
Similar to the previous experiment, we train AutoBot on each city and evaluate it on the rest of cities. We employed nuPlan \cite{caesar2021nuplan} data for this experiment due to the large number of samples existing in diverse cities and selected $10$K samples from each city.
The results are shown in \Cref{tab:multi_city_evaluation}.
It shows that the performance of AutoBot drops once evaluated on other cities.
For instance, the first row shows that the model trained on Pittsburgh has the best performance on Pittsburgh (brier-minFDE $2.4$) and worse performances on Boston ($2.7$) and Singapore ($3.5$). 
This indicates a clear generalization gap between cities, emphasizing the discrepancies between different environments. 
Moreover, it can also be observed that the model trained on Singapore performs the worst on average.
This is an expected outcome, given that Singapore is a left-hand traffic city, unlike the other ones.

\textbf{Takeaways: } The findings in this section 
reveal that \textit{state-of-the-art models trained on recent large-scale datasets struggle to generalize to new domains}. As a concrete recommendation, it highlights the importance of geographical diversity in the data collection process for both training and evaluation.


\begin{table}
\vspace{-0.3cm}
\centering
\begin{minipage}{0.6\textwidth}
\begin{tabular}{@{}l*{4}{c}@{}}
\toprule
& \multicolumn{4}{c}{\cellcolor{orange!25} $\leftarrow$ \textbf{Evaluation} $\rightarrow$} \\
\cellcolor{blue!15}{$\downarrow$ \textbf{Training}} & \cellcolor{orange!25} {Pittsburgh} & \cellcolor{orange!25} {Boston} &  \cellcolor{orange!25} {Singapore} & \cellcolor{orange!25} \textit{Average} \\
\midrule
\cellcolor{blue!15} Pittsburgh & \textbf{2.4} & 2.7 & 3.5 & \textbf{2.8}\\
\cellcolor{blue!15} Boston & 4.1 & \textbf{2.2} & 3.4 & 3.2 \\
\cellcolor{blue!15} Singapore & 4.9 & 3.5 & \textbf{2.1} & 3.5\\
\bottomrule
\end{tabular}
\end{minipage}
\hfill
\begin{minipage}{.35\textwidth}
\centering
\caption{\textbf{Cross-city generalization experiment.} We train and validate AutoBot across multiple cities in the nuPlan dataset and report the brier-minFDE ($\downarrow$) metric.
\label{tab:multi_city_evaluation}
}
\end{minipage}
\vspace{-0.3cm}
\end{table}

\subsection{Scaling data to 2M trajectories.}    
\label{sec:multi_dataset_training}
\begin{figure}[htbp]
  \centering
  \begin{minipage}{0.45\textwidth}
    \centering
\captionof{table}{\textbf{nuScenes Leaderboard.} We train AutoBot and MTR with all datasets, and evaluate on nuScenes (ranking at the time of submission among public methods)
}
\centering
  \begin{tabular}{@{}l@{}c@{}c@{}}
    \toprule
    
Method   & Ranking ($\downarrow$)& minADE$5$ ($\downarrow$)  \\
\midrule
\textbf{MTR-UniTraj}             & 1  & 0.96   \\
Goal-LBP \cite{Goal_LBP}  & 2  & 1.02  \\
CASPNet++ \cite{schäfer2023caspnet} & 3  & 1.16 \\
Socialea \cite{chen2023q} & 4 & 1.18  \\
Autobot-Unitraj         & 11 & 1.26 \\
Autobot                 & 19 & 1.37  \\
 \bottomrule
\end{tabular}
\label{tab:nuscenes_leaderboard}
 \end{minipage}
\hfill
 \begin{minipage}{0.48\textwidth}
    \centering
    \captionof{figure}{\textbf{Relationship between dataset size and model performance}. The prediction error of AutoBot as the combined dataset size increases, varying from 20\% to 100\% of the total data. }
    \includegraphics[width=0.9\linewidth]{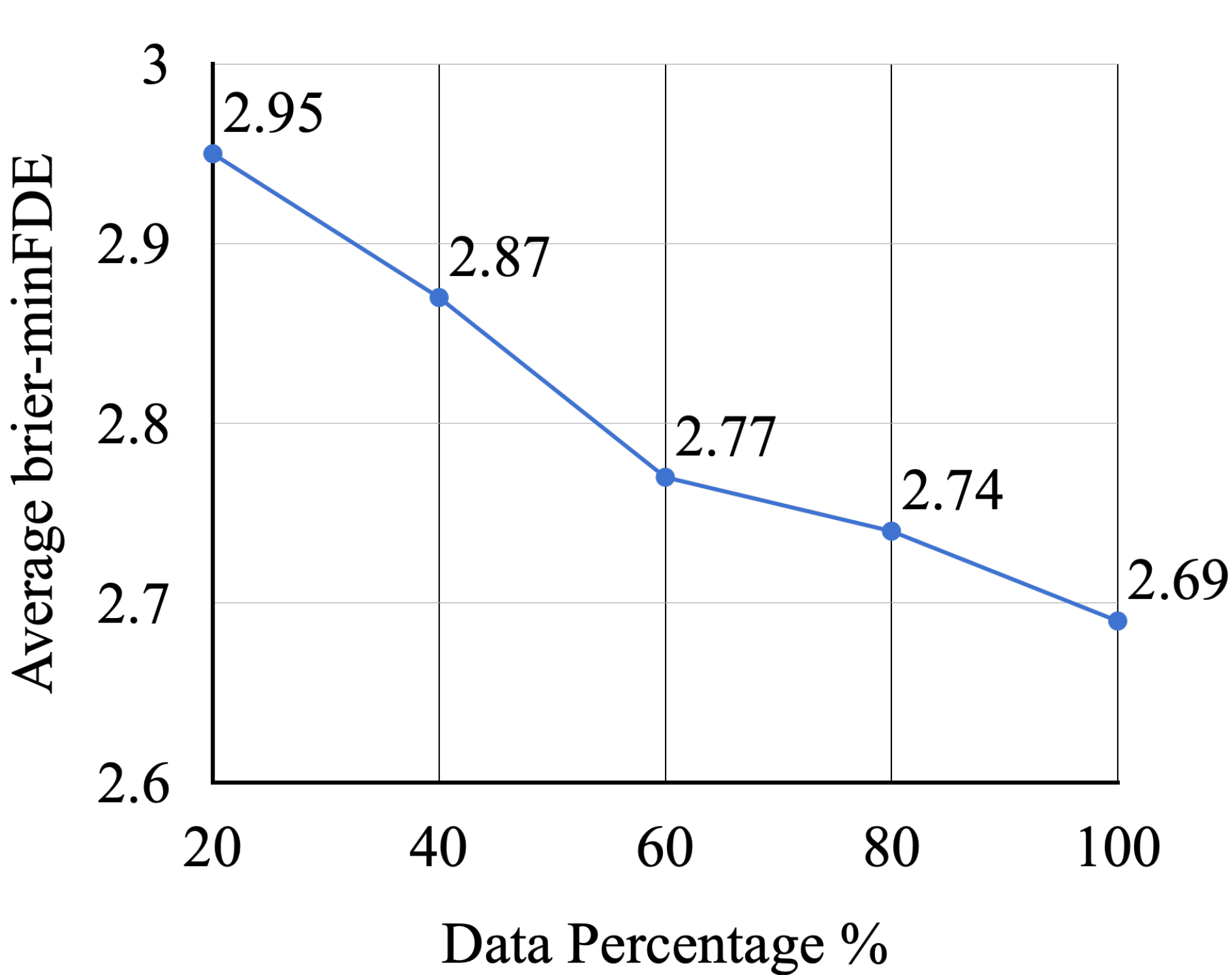}
\label{fig:data_percentage}
    \end{minipage}
\end{figure}

The unified data available in UniTraj forms the largest public data one can use to train a trajectory prediction model. In this section, we explore if we can improve the models' performance by simply scaling the size of the training dataset.
Therefore, we combine all the existing real datasets in UniTraj into a single large training set on which we train the considered models. 
The results of this experiment are presented at the bottom of \Cref{tab:cross_dataset_evaluation} (row `All'), demonstrating improvements over the model trained solely on a single dataset. While the improvements are not identical for different datasets, they are particularly significant in the case of the nuScenes dataset, making the MTR model trained on combined data rank $1$\textsuperscript{st} in the nuScenes leaderboard (shown in \Cref{tab:nuscenes_leaderboard}). 
For instance, training the MTR model on "All" datasets enables it to outperform the model trained on nuScenes and Argoverse2 by a large margin.
Moreover, while the performance on \waymo{} has not been improved, the resulting model performs much better than the model trained on \waymo{} on other datasets. 
These improvements are attributed to the relatively larger size and diversity of the combined dataset compared to each individual one. We elaborate on this more in \Cref{sec:interpreting}.

The table also shows that certain models benefit more from larger data sizes compared to others. Specifically, when looking at the performance on Argoverse~2 and nuScenes datasets, which benefit from the increased dataset size, models like MTR and Wayformer show more significant improvements than AutoBot. This difference in performance enhancement is attributed to the models' capacity. For instance, MTR has $60.1$ M parameters, providing it with a higher capacity to learn from larger datasets, whereas AutoBot, with only $1.5$ M parameters, may not be as able to utilize the additional data. 

In order to illustrate the impact of data size on the performance of a trajectory predcition model, we gradually increase the number of training samples from $20$\% to $100$\% of the combined dataset. We then report the AutoBot model's performance using the average brier-minFDE metric among all three datasets. \Cref{fig:data_percentage} shows the curve revealing a consistent reduction in the prediction error once the dataset size increases.
This highlights the substantial benefits of larger datasets on the model's performance and offers prospects for improved performances with larger data sizes. 

\textbf{Takeaways:} The experimental results underscore the potential and need for larger, more diverse datasets in the trajectory prediction field. Such datasets will also push the boundaries of the current performances of the models. 

\subsection{Analyzing the results}
\label{sec:interpreting}
In this section, we aim to explain the findings about generalization gaps and the data scaling impact in \Cref{sec:cross_dataset_evaluation,sec:multi_dataset_training}.
Thus, we first delve into a comparative analysis of datasets integrated within the UniTraj framework with our fine-grained evaluations. We then employ these insights to explain the findings. 

\subsubsection{Dataset analysis.}
\label{sec:dataset_analysis}
We provide an in-depth comparison between datasets to help the understanding of the results presented in previous subsections. Moreover, the insights aid in making informed decisions about selecting the most appropriate datasets and settings for specific research or application needs.

\input{figs/fine_grained_eval}

\noindent\textbf{Trajectory type based comparison:}
The analysis of trajectory types in the \waymo{}, Argoverse~2, and nuScenes datasets in \Cref{fig:traj_type} reveals trajectory type imbalances, primarily featuring a prevalence of straight trajectories, constituting 54\% to 68\% of all trajectories, with minimal instances of u-turns.
\waymo{} exhibits a notably diverse trajectory mix, with a significant number of left and right turn trajectories, approximately two times more than what is observed in Argoverse~2 and nuScenes. This stands in contrast to Argoverse~2 and nuScenes, which primarily contain straight trajectories over varied turning maneuvers.

\noindent\textbf{Kalman difficulty based comparison:}
The distribution of sample difficulties within the \waymo{}, Argoverse~2, and nuScenes datasets, shown in \Cref{fig:kalman}, exhibits a consistent trend where easier scenarios significantly outnumber more challenging ones.
\waymo{} demonstrates a relatively balanced distribution across lower to moderate difficulty levels, with around $\sim$24\% of trajectories falling within the easiest category (Kalman difficulties up to 10), with a consistent presence observed up to a Kalman difficulty of 50.0.
In contrast, both nuScenes and Argoverse~2 exhibit a substantial bias towards easier difficulties, comprising approximately $\sim$42\% and $\sim$46\% of samples, respectively, in the lowest difficulty range (Kalman difficulties lower than 10), and show a sharp decrease in proportion with increasing difficulty levels, indicating datasets primarily composed of simpler scenarios which may potentially limit their efficacy in training models for more complex situations.

\begin{table*}
\caption{\textbf{Stratified evaluations per trajectory type.} We report brier-minFDE on nuScenes validation data. We compare the performance of two MTR \cite{shi2022motion_mtr} models trained on nuScenes data (nuScenes) and the combined dataset in UniTraj (All).}
\centering
\resizebox{\textwidth}{!}{
\begin{tabular}{l c c c c c c c c c}
\toprule
\text{Traj. Type} & \text{Stationary} & \text{Straight} & \text{Straight right} & \text{Straight left} & \text{Right u-turn} & \text{Right-turn} & \text{Left u-turn}& \text{Left-turn} & \text{All}\\
\midrule
\text{MTR (nuScenes)}  & 2.15 & 2.58 & 4.85 & 4.26 & 8.13 & 4.82 & 5.17 & 4.85 & 2.86\\
\text{MTR (All)}     & 2.23 & 2.31 & 3.13 & 3.06 & 2.98 & 3.53 & 2.10 & 2.82 & 2.27\\
\bottomrule
\end{tabular}}

\label{tab:Profiling_Finegrained_evaluation_Trajectory_type}
\end{table*}

\begin{table}[ht]
\centering
\begin{minipage}{.49\textwidth}
\caption{\textbf{Cross-dataset generalization experiments with identical sample size.} We select $30$K random samples from every dataset, then train and validate AutoBot across them and report the brier-minFDE metric.}
\centering
\resizebox{\textwidth}{!}{
\begin{tabular}{lcccc}
\toprule
& \multicolumn{4}{c}{\cellcolor{orange!25} {$\leftarrow$ \textbf{Evaluation} $\rightarrow$}} \\
\cellcolor{blue!15} $\downarrow$ \textbf{Training} & \cellcolor{orange!25} {nuScenes} & \cellcolor{orange!25} {Argoverse 2} & \cellcolor{orange!25} {\waymo{}} & \cellcolor{orange!25} {Average} \\
\midrule
\cellcolor{blue!15} nuScenes & \textbf{3.38} & 4.48 & 6.88 & 4.91 \\
\cellcolor{blue!15} Argoverse 2 & 4.67 & \textbf{2.90} & 5.07 & 4.21 \\
\cellcolor{blue!15} \waymo{} & 4.42 & 4.04 & \textbf{3.22} & \textbf{3.89} \\
\midrule
\cellcolor{blue!15} All & \textbf{3.25} & \textbf{2.80} & \textbf{3.13} & \textbf{3.06} \\
\bottomrule
\end{tabular}}
\label{tab:cross_dataset_evaluation_30k}
\end{minipage}
\begin{minipage}{.49\textwidth}
\caption{\textbf{Fine-grained evaluation Kalman difficulty.} We report the brier-minFDE ($\downarrow$) metric across three chunks of Kalman difficulties on nuScenes validation data. We compare the performance of two MTR \cite{shi2022motion_mtr} models trained on nuScenes data (nuScenes) and the combined dataset in UniTraj (All).}
\centering 
\resizebox{\textwidth}{!}{
\begin{tabular}{lccc}
\toprule
{Kalman } & {Easy} & {Medium} & {Hard} \\
 difficulty & $\in [0, 30[$ & $\in [30, 50[$ & $\in [50, 100[$ \\
\midrule
MTR (nuScenes) & 2.73 & 4.52 & 4.25 \\
MTR (All)      & 2.23 & 2.97 & 4.20 \\
\bottomrule
\end{tabular}}
\label{tab:Profiling_Finegrained_evaluation_Kalman}
\end{minipage}
\end{table}

\noindent\textbf{Explaining the findings in cross-dataset generalization 
and multi-dataset training experiments}:
Our cross-dataset generalization experiment in \Cref{sec:cross_dataset_evaluation} shows that models do not generalize equally across different datasets. Notably, models trained on \waymo{} generalize better to other datasets. Furthermore, models trained on combined datasets exhibit considerable improvements.
To understand these phenomena, it's important to delve into the differences between datasets, focusing on two main aspects: size and diversity. 

To investigate the impact of dataset size, we replicate the cross-dataset generalization experiments (\Cref{tab:cross_dataset_evaluation}), but with control on the dataset size, as we select $30$k random samples for each dataset's training set. \Cref{tab:cross_dataset_evaluation_30k} shows the results. 
The last column shows the generalization hierarchy where again the \waymo{} generalizes best, followed by Argoverse~2 and then nuScenes. This shows that better cross-dataset generalization is not solely attributed to the size of the datasets. The last row illustrates that for multi-dataset training, there is a considerable improvement for all the datasets, highlighting the pronounced benefit of adding more data in small-scale dataset scenarios. 

\Cref{sec:dataset_analysis} reveals that the datasets are dissimilar in terms of diversity. Notably, \waymo{} encompasses the most diverse range of scenarios in comparison to other datasets. This explains the superior generalization of \waymo{} to other datasets, as 
the diversity enables models to learn the full spectrum of data distributions more comprehensively. Similarly, the combined dataset provides a more diverse collection of trajectories, leading to enhanced performance for the models. 
To demonstrate this, we compare the fine-grained evaluations of MTR model trained on nuScenes and the combined dataset. 
\Cref{tab:Profiling_Finegrained_evaluation_Trajectory_type} shows the per trajectory type performance of the two models where the model trained on full data outperforms in every trajectory type since the full data includes significantly more samples from each trajectory type. We also compare the performances using the Kalman difficulty measure in \Cref{tab:Profiling_Finegrained_evaluation_Kalman}. The combined data has considerably more medium-difficulty samples (shown in \Cref{fig:kalman}) leading to significant performance improvements in the medium-range samples. These results highlight the importance of diversity in the data.

\section{Conclusions}
In conclusion, our study examines two critical research questions essential for advancing the field of vehicle trajectory prediction. We have uncovered that models face significant challenges in generalizing across different domains (RQ1), exhibiting considerable performance drops when encountering new datasets or cities. Additionally, our findings affirm that larger, more diverse datasets significantly boost model performance and generalization capabilities (RQ2), underscoring the importance of data richness. Besides, we release the UniTraj framework as a versatile tool that opens up new opportunities for exploration in trajectory predictions. 
We believe that this framework will help significantly in advancing 
research in the field of trajectory prediction.


\vspace{2mm}
\noindent\textbf{Acknowledgments:} Authors would like to thank Mickaël Chen, Quanyi Li, Ahmad Rahimi, Yihong Xu, Olaf Dunkel and Ahmad Salimi for their valuable discussions and help on preliminary versions of this work. The project was partially funded by Honda R\&D Co., Ltd, Valeo Paris, and ANR grant MultiTrans (ANR-21-CE23-0032).

{\small
\bibliographystyle{ieee_fullname}
\bibliography{egbib,other_ref}
}


\newpage
\section*{Appendix}
\appendix

\input{appendix.tex}
\end{document}

%% file: figs/fine_grained_eval.tex
\begin{figure}[t]
\begin{subfigure}{0.49\linewidth}
\vspace{0pt} 
\trimbox{0.3cm 0cm 0cm 0cm}{ 
\begin{tikzpicture}
  \begin{axis}[
    ybar=.3pt,
    width=1.08\linewidth,
    height=5cm,
    bar width=4pt,
    ylabel={\# trajectories (\%)},
    ylabel style={yshift=-7pt, align=center},
    ytick pos=left,
    xtick pos=bottom,
    xtick=data,
    xticklabels={Stationary, Straight, Straight Right, Straight Left, Right Turn, Left Turn, Left U-Turn, Right U-Turn},
    every x tick label/.append style={font=\scriptsize, rotate=45, anchor=east}, 
    ymin=0,
    legend style={
      at={(1., 1.)},
      draw=none,
      anchor=north east,
      cells={anchor=west},
      legend columns=1,
      font=\scriptsize,
    },
    enlarge x limits={abs=0.45},
    legend image code/.code={%
        \draw[#1] (0cm,-0.1cm) rectangle (4pt,5pt);},
    ]
    
    \draw[dotted] (axis cs:\pgfkeysvalueof{/pgfplots/xmin}, 20) -- (axis cs:\pgfkeysvalueof{/pgfplots/xmax}, 20);
    \draw[dotted] (axis cs:\pgfkeysvalueof{/pgfplots/xmin}, 40) -- (axis cs:\pgfkeysvalueof{/pgfplots/xmax}, 40);
    \draw[dotted] (axis cs:\pgfkeysvalueof{/pgfplots/xmin}, 60) -- (axis cs:\pgfkeysvalueof{/pgfplots/xmax}, 60);
    
    \addplot[fill=Blue!75,draw=Blue!30!black] coordinates {
          (0, 3.2596956361743965)
          (1, 54.013746235734985)
          (2, 3.7757510566071426)
          (3, 4.283858788075419)
          (4, 14.849766904081193)
          (5, 19.422518742392537)
          (6, 0.36325293465693204)
          (7, 0.03140970227739394)
    };
%
    \addplot[fill=ForestGreen!75,draw=ForestGreen!30!black] coordinates {
          (0, 7.055467373577043)
          (1, 67.98394178898506)
          (2, 3.366551575548319)
          (3, 2.938914434389881)
          (4, 8.371106129283872)
          (5, 10.182563968298124)
          (6, 0.0692728532233695)
          (7, 0.032181876694321265)
    };
%
    \addplot[fill=GreenYellow!75,draw=GreenYellow!30!black] coordinates {
          (0, 2.738719832109129)
          (1, 70.05596362364463)
          (2, 4.0573627142357465)
          (3, 4.41063308849248)
          (4, 8.719832109129067)
          (5, 9.828611402588319)
          (6, 0.1259181532004197)
          (7, 0.06295907660020986)
    };

    \legend{\waymo{}, Argoverse 2, nuScenes}
  \end{axis}
\end{tikzpicture}
}
\caption{\textbf{Distribution of trajectory types across datasets.}}
\label{fig:traj_type}
\end{subfigure}
\hfill
%
\begin{subfigure}{0.48\linewidth}
\vspace{0pt} 
\centering
\trimbox{0.2cm 0cm 0cm 0cm}{ 
\begin{tikzpicture}
  \begin{axis}[
    ybar=.3pt,
    width=1.14\linewidth,
    height=5cm,
    bar width=4pt,
    ylabel={\# trajectories (\%)},
    ylabel style={yshift=-7pt, align=center},
    xlabel={Kalman difficulty},
    xlabel style={yshift=4pt, align=center},
    ytick pos=left,
    xtick pos=bottom,
    xtick=data,
    xticklabels={0--10, 10--20, 20--30, 30--40, 40--50, 50--60, 60--70, 70--80}, 
    every x tick label/.append style={font=\footnotesize, rotate=45, anchor=east}, 
    ymin=0,
    legend style={
      at={(1., 1.)},
      draw=none,
      anchor=north east,
      cells={anchor=west},
      legend columns=1,
      font=\scriptsize,
    },
    enlarge x limits={abs=0.45},
    legend image code/.code={%
        \draw[#1] (0cm,-0.1cm) rectangle (4pt,5pt);},
    ]
    
    \draw[dotted] (axis cs:\pgfkeysvalueof{/pgfplots/xmin}, 10) -- (axis cs:\pgfkeysvalueof{/pgfplots/xmax}, 10);
    \draw[dotted] (axis cs:\pgfkeysvalueof{/pgfplots/xmin}, 20) -- (axis cs:\pgfkeysvalueof{/pgfplots/xmax}, 20);
    \draw[dotted] (axis cs:\pgfkeysvalueof{/pgfplots/xmin}, 30) -- (axis cs:\pgfkeysvalueof{/pgfplots/xmax}, 30);
    \draw[dotted] (axis cs:\pgfkeysvalueof{/pgfplots/xmin}, 40) -- (axis cs:\pgfkeysvalueof{/pgfplots/xmax}, 40);

    \addplot[fill=Blue!75,draw=Blue!30!black] coordinates {
        (0, 23.952629169758517)
        (1, 24.611144221251667)
        (2, 21.142773679220554)
        (3, 15.996653629820514)
        (4, 9.446714336574939)
        (5, 3.7812939732733786)
        (6, 0.8821063091988663)
        (7, 0.1533644322320915)
    };

     \addplot[fill=ForestGreen!75,draw=ForestGreen!30!black] coordinates {
        (0, 42.29571326493321)
        (1, 27.38459524471863)
        (2, 18.436942612622932)
        (3, 8.649833908788924)
        (4, 2.555459191743985)
        (5, 0.5563646479357236)
        (6, 0.10200018545488264)
        (7, 0.01581821057856469)
    };

    \addplot[fill=GreenYellow!75,draw=GreenYellow!30!black] coordinates {
        (0, 46.08954179783141)
        (1, 30.622595313046517)
        (2, 15.928646379853095)
        (3, 5.771248688352571)
        (4, 1.3326337880377754)
        (5, 0.24484085344526058)
        (6, 0.006995452955578873)
        (7, 0.0034977264777894365)
    };

    \legend{\waymo{}, Argo.\ 2, nuScenes}
    
  \draw[black,->] (0.9,43) -- (0.4,43) node[midway, right]{};
  \begin{scope}
    \node[anchor=south west,inner sep=0,draw=black,line width=1pt] at (0.6,39) {\includegraphics[width=2.76cm]{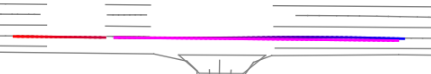}};
  \end{scope}
  
  \draw[black,->] (2,30) -- (1.4,30) node[midway, right]{};
  \begin{scope}
    \node[anchor=south west,inner sep=0,draw=black,line width=1pt] at (2.1,21.4) {\includegraphics[width=2.8cm]{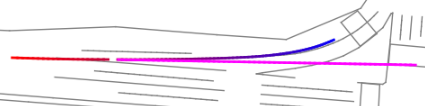}};
  \end{scope}
  
  \draw[black,->] (4.2,18.5) -- (2.4,18.5) node[midway, right]{};
  \begin{scope}
    \node[anchor=south west,inner sep=0,draw=black,line width=1pt] at (4.3,5) {\includegraphics[width=1.5cm]{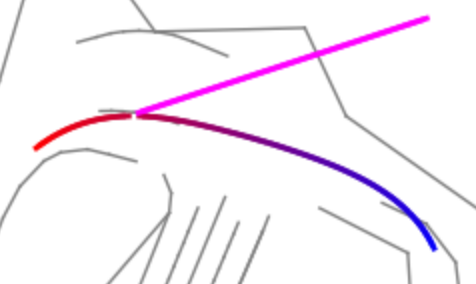}};
  \end{scope}
  
  \end{axis}
  
\end{tikzpicture}
}
\caption{\textbf{Histogram of the Kalman Difficulty of trajectories.}} 
\label{fig:kalman}
\end{subfigure}
\caption{
Figure (a) shows the distribution of trajectory types. It reveals an imbalance across different types with straight being the most common trajectory type in the datasets.
Figure (b) shows the histogram of the Kalman Difficulty of trajectories. To give a sense of the Kalman difficulty, we overlay three random examples. The past trajectory, the ground truth, and the Kalman filter prediction are shown in red, blue, and magenta, respectively. 
The plot displays a clear trend with a notably higher count of simpler scenarios compared to challenging ones. \waymo{}, in particular, shows a relatively balanced distribution across scenarios.
}
\end{figure}

%% file: appendix.tex
In this supplementary document, we provide additional content to complement our main paper. This includes a set of experiments that further illuminate the capabilities of the UniTraj framework for the research community (see \Cref{sec:appendix_further_exps}), additional results including cross-checking the performance of models and reporting other metrics (see \Cref{sec:appendix_complementory_results}), and more details about UniTraj framework (see \Cref{sec:appendix_additional_details}). 

\section{Further experimentation with UniTraj}
\label{sec:appendix_further_exps}
UniTraj is a flexible and comprehensive tool that opens up various research opportunities. While we have shown some of these in the main paper, we demonstrate other opportunities and capabilities of the framework here. 

\subsection{Continual learning on multiple datasets} 
\label{sec:appendix_forgetting}
Adapting trajectory prediction models to new datasets without erasing prior knowledge, termed catastrophic forgetting \cite{kirkpatrick2017overcoming}, is crucial for autonomous driving models.
Our results in \Cref{tab:continual_learning} illustrate this challenge, where we train AutoBot on $30$k samples of \waymo{} (resp.\ Argoverse~2) dataset followed by fine-tuning on Argoverse~2 (resp.\ \waymo{}).
Fine-tuning a \waymo{}-trained model with Argoverse~2 data modestly improves performance in Argoverse~2 (improvement +0.92), but significantly hurts its \waymo{} performance (-0.52).
Similarly, fine-tuning an Argoverse~2-trained model with \waymo{} data notably boosts \waymo{} performance (+2.06) but severely impacts Argoverse~2 performance (-0.98). 
This is attributed to the considerable domain gap between \waymo{} and Argoverse~2, as shown in our cross-dataset generalization experiment. This outcome exemplifies the issue of catastrophic forgetting, highlighting the challenge of balancing adaptation to new datasets without compromising performance on previously learned ones.

\begin{table}
  \centering
    \caption{\textbf{Continual learning results.} This table presents outcomes from two methodologies: initial training on \waymo{} with subsequent fine-tuning on Argoverse~2, and initial training on Argoverse~2 followed by fine-tuning on \waymo{}. We report the brier-minFDE metric for the validation sets. The "Improvement" column indicates performance gain on the fine-tuning domain data, and the "Forgetting" column shows the performance drop on the training domain data.}
  \begin{tabular}{@{}l@{\hspace{.1cm}}c@{\hspace{.1cm}}lcc@{}}
    \toprule
    
    Training & \multirow{2}{*}{$\rightarrow$} & Finetuning & Improvement & Forgetting \\
    domain & & domain & (finetun. dom.) & (train. dom.) \\
    \midrule
    \waymo{} & $\rightarrow$ & Argoverse~2 &  + 0.92 & - 0.52 \\
    Argoverse~2 & $\rightarrow$ & \waymo{} & + 2.06 & - 0.98 \\
    \bottomrule
  \end{tabular}

  \label{tab:continual_learning}
\end{table}

\begin{table}[t]
\centering
\caption{\textbf{Generalization from synthetic data to real data.} We train AutoBot with nuScenes or/and synthetic dataset, and evaluate on nuScenes.}
  \begin{tabular}{@{}l@{\hspace{.4cm}}c@{\hspace{.4cm}}c@{}}
    \toprule
    
Training       & minADE6 ($\downarrow$) & brier-minFDE ($\downarrow$) \\
\midrule
PG                & 8.36 & 14.35    \\
nuScenes          & \textbf{1.28} & \textbf{3.55}     \\
nuscenes and PG & 1.30 & 3.56    \\
 \bottomrule
\end{tabular}

\label{tab:synthetic_dataset}
\end{table}

\subsection{Synthetic data}
\label{sec:appendix_synthetic_data}
Synthetic data serves as a cost-effective and versatile approach, particularly in autonomous driving, thanks to its ease of generation and capacity to simulate various, even rare, driving scenarios during training.
In our approach, we leverage a large synthetic dataset created through Procedural Generation (PG) to pre-train our models. PG, a technique used in the MetaDrive simulator~\cite{li2022metadrive}, facilitates the creation of varied traffic scenarios and maps based on predefined rules. In our experiment, we use the same setting of MetaDrive simulator; traffic density at 15 vehicles per 100 meters and incorporating 2 roadblocks in each scenario. This resulted in the creation of 30,000 unique scenarios.

We train AutoBot using the synthetic PG dataset and subsequently fine-tune it on the nuScenes dataset.
The pre-training stage can be helpful for injecting basic behaviors in models, such as lane following and avoiding collisions.
\Cref{tab:synthetic_dataset} shows the result of our experiment. 
The comparison between the first and second rows of the table reveals a notable domain gap between datasets. Specifically, the model trained on the PG dataset underperforms significantly on the nuScenes dataset. Moreover, the model does not show any substantial performance improvement after fine-tuning, compared to those that have not undergone pre-training. 
This outcome indicates that while synthetic data may potentially serve as a useful starting point for training and introducing models to a range of conditions, its effectiveness in improving real-world performance may be limited.
This limitation is attributed to the domain gap between real and synthetic data, as well as the simplicity and limited diversity of the synthetic data.
We believe that more realistic and diverse synthetic datasets might be helpful as a pre-training source.

\begin{table*}
\caption{\textbf{Stratified evaluations per trajectory type.} We report brier-minFDE for AutoBot \cite{girgis2022autobot} and MTR \cite{shi2022motion_mtr} models trained and evaluated on \waymo{} dataset.}
\centering
\resizebox{\textwidth}{!}{
\begin{tabular}{l c c c c c c c c c}
\toprule
\text{Traj. Type} & \text{Stationary} & \text{Straight} & \text{Straight right} & \text{Straight left} & \text{Right u-turn} & \text{Right-turn} & \text{Left u-turn}& \text{Left-turn} & \text{All}\\
\midrule
\text{AutoBot} 
& 1.50 &
2.21 & 2.77 & 2.69 & 8.06 & 2.99 & 4.32 & 2.69 & 2.47 \\
\text{MTR}
& 1.09 & 2.13 & 2.86 & 2.90 & 5.96 & 2.83 & 4.58 & 2.64 & 2.12\\
\bottomrule
\end{tabular}}

\label{tab:Finegrained_evaluation_Trajectory_type}
\end{table*}

\begin{table}
\caption{\textbf{Fine-grained evaluation Kalman difficulty.} We report the brier-minFDE metric across three chunks of Kalman difficulties. We compare AutoBot \cite{girgis2022autobot} and MTR \cite{shi2022motion_mtr} models trained and evaluated on \waymo{} dataset. 
}
\[
\begin{array}{@{}l*{5}{c}@{}}
\toprule
\text{Kalman} & \text{Easy} & \text{Medium} & \text{Hard} & \text{All} \\
\text{difficulty} & \in \text{$\left [0, 30\right [$} & \in \text{$\left [  30, 60\right [$} & \in \text{$\left [  60, 100\right [$} & \\
\midrule
\text{AutoBot}
& 2.52 & 2.46 & 3.52  & 2.47\\
\text{MTR }
& 2.05 & 2.40 & 2.45 & 2.12 \\
\bottomrule
\end{array}
\]
\label{tab:Finegrained_evaluation_Kalman}
\end{table}

\subsection{Comparing different models' performances using fine-grained evaluations:}
We compare the performance of multiple trajectory prediction models in Table 2 of the main paper.  
The table provides a comparison between different models. Specifically, it shows performances of AutoBot and MTR on the \waymo{} dataset, where MTR achieves a lower error rate of $2.37$, outperforming AutoBot's $2.47$.
However, this is a course comparison without any details about the strengths and weaknesses of each model. 
Thanks to the fine-grained evaluation approach available in UniTraj, we provide a more detailed comparative analysis in \Cref{tab:Finegrained_evaluation_Trajectory_type,tab:Finegrained_evaluation_Kalman}. The first table compares performances across different trajectory types. While MTR generally outperforms AutoBot, it falls behind in specific trajectories such as straight right, straight left, and left U-turns.  The second table also reveals that AutoBot exceeds MTR in medium difficulty scenarios. These insights offer a more thorough comparison and help pinpoint specific areas of weakness in the models.

\subsection{Other potential future directions}
UniTraj framework paves the way toward building foundation models for trajectory forecasting. Foundation models are typically trained on vast datasets, which, as of now, are not readily available in the trajectory forecasting domain. An effective workaround is to utilize extensive synthetic data for initial model training, followed by fine-tuning on real-world datasets.
UniTraj facilitates this approach by enabling the integration of various synthetic datasets alongside its collection of the largest real data currently available in this field. This direction of research is already gaining traction and some recent studies in trajectory forecasting have adopted the foundation model concept by tokenizing the action space and focusing on predicting the subsequent token \cite{seff2023motionlm,philion2023trajeglish}. 

The framework also opens up opportunities for other research studies such as coreset selection and dataset distillation \cite{guo2022deepcore,yu2023distillation}. This involves creating a compact subset of the trajectory data that encapsulates the majority of information from the combined dataset. 

\section{Complementary results}
\label{sec:appendix_complementory_results}
In this section, we present additional results that complement those featured in the main paper. We first provide performances of the prediction models integrated into the framework compared with their officially reported performance to verify our correct integration. Then, we report metrics beyond brie-minFDE for the results we presented in the paper. 

\subsection{Cross-checking the performance of baselines with original papers}

For MTR, we report numbers on \waymo{}. We trained MTR on \waymo{} and in the same setting as the official setting, e.g., 1 second of past trajectories and 8 seconds of future trajectories. Then, we evaluate the model on the official \waymo{} validation set, using the official evaluation tool to report the numbers. 
The results in \Cref{tab:MTR} show that the integrated MTR can achieve similar performance compared to the original implementation.
\begin{table}
\caption{\textbf{Performance evaluation of MTR in the \waymo{} setting compared to the original implementation.}
}
\label{tab:MTR}
\centering
\begin{tabular}{@{}ccccc@{}}
\toprule
             & mAP  & minADE & minFDE & MissRate \\ \midrule
Vehicle      & 0.44 & 0.78   & 1.55   & 0.16     \\
Pedestrian   & 0.43 & 0.35   & 0.73   & 0.07     \\
Cyclist      & 0.36 & 0.72   & 1.45   & 0.19     \\
Avg (ours)     & 0.41 & 0.62   & 1.24   & 0.14     \\
\midrule
Avg-original & 0.42 & 0.60   & 1.23   & 0.14     \\ \bottomrule
\end{tabular}
\end{table}

Similarly, for Wayformer, we report numbers on \waymo{} with the same data setting as the official setting. 
Since the original paper has not reported numbers on validation set, we compare our performance with MTR.
The results in \Cref{tab:wayformer} show that our internal implementation of Wayformer outperforms MTR in terms of  minADE and minFDE metrics. However, due to 
the absence of certain details from the Wayformer's non-public implementation, there remains a discrepancy in mAP metric performance.

\begin{table}[]
\caption{
\textbf{Performance evaluation of Wayformer in the \waymo{} setting.}
}
\centering
\label{tab:wayformer}
\begin{tabular}{@{}ccccc@{}}
\toprule
           & mAP  & minADE & minFDE & MissRate \\ \midrule
Vehicle    & 0.28 & 0.67   & 1.39   & 0.14     \\
Pedestrian & 0.25 & 0.32   & 0.67   & 0.09     \\
Cyclist    & 0.24 & 0.68   & 1.40   & 0.21     \\
Avg (ours)        & 0.26 & 0.56   & 1.15   & 0.15     \\ 

\bottomrule
\end{tabular}
\end{table}

For Autobot, please refer to our results on the nuScenes leaderboard in in Table 4 in the main paper.


    



\subsection{Reporting other metrics}
\label{sec:appendix_other_metrics}
We extend our examination of model generalization capabilities, previously reported in the main paper using the brierFDE metric (Table~2), by presenting additional evaluation metrics: minFDE in \Cref{tab:cross_dataset_evaluation_minFDE}, minADE in \Cref{tab:cross_dataset_evaluation_minADE} and MR in \Cref{tab:cross_dataset_evaluation_MR}. We report the metrics of the same experiments; training models on each dataset individually and evaluating their performance across all others.

Our expanded analysis confirms the observed trend of reduced model performance on unfamiliar datasets, a consistent result across the considered model architectures and datasets. It also reinforces the conclusions drawn in the main paper, illustrating the generalization hierarchy where \waymo{}-trained models outperform others, followed by Argoverse~2 and nuScenes.

We further confirm the benefit of training models on combined datasets, which aligns with the main paper's insights, showing notable performance improvements, especially for nuScenes. 

\begin{table*}
\caption{\textbf{Cross-dataset generalization and multi-dataset training experiments minFDE metric.}
}
\resizebox{1.0\textwidth}{!}{
\begin{tabular}{@{}l r c *{8}{@{\hspace{0.05cm}}c} @{}}
\toprule
& & \multicolumn{3}{c}{\text{MTR \cite{shi2022motion_mtr}}} & \multicolumn{3}{c}{\text{Wayformer * \cite{nayakanti2023wayformer}}} & \multicolumn{3}{c}{\text{AutoBot \cite{girgis2022autobot}}} \\
\cmidrule(lr){3-5} \cmidrule(lr){6-8} \cmidrule(lr){9-11}
& & \multicolumn{9}{c}{\cellcolor{orange!25} $\leftarrow$ \textbf{Evaluation} $\rightarrow$}\\
\cellcolor{blue!15}{$\downarrow$ \textbf{Training}} & \#\text{trajs} & \cellcolor{orange!25} \text{nuScenes}  & \cellcolor{orange!25} \text{Argoverse~2} & \cellcolor{orange!25} \text{\waymo{}} & \cellcolor{orange!25} \text{nuScenes}  & \cellcolor{orange!25} \text{Argoverse~2} & \cellcolor{orange!25} \text{\waymo{}} & \cellcolor{orange!25} \text{nuScenes}  & \cellcolor{orange!25} \text{Argoverse~2} & \cellcolor{orange!25} \text{\waymo{}}\hspace{-0.3cm}\\
\midrule
\cellcolor{blue!15} \text{nuScenes} & 32\text{k} & \textbf{2.33} & 3.89 & \multicolumn{1}{c|}{6.72}  & \textbf{2.50} & 3.93 & \multicolumn{1}{c|}{6.48} & \textbf{2.62} & 3.70 & 5.85 \\
\cellcolor{blue!15} \text{Argoverse~2} & 180\text{k} & 3.10 & \textbf{1.68} & \multicolumn{1}{c|}{4.04} & 3.07 & \textbf{1.75} & \multicolumn{1}{c|}{4.12} & 3.52 & \textbf{1.70} & 3.59 \\
\cellcolor{blue!15} \text{\waymo{}  } & 1800\text{k} & 2.52 & 3.14 & \multicolumn{1}{c|}{\textbf{1.78}} & 2.51 & 3.14 & \multicolumn{1}{c|}{\textbf{1.46}}  & 2.90 & 2.41 & \textbf{1.65} \\
\midrule
\cellcolor{blue!15} \text{All} &  2012\text{k} & \textbf{1.81} & \textbf{1.61} & \multicolumn{1}{c|}{\textbf{1.78}}  & \textbf{1.76} & \textbf{1.51} & \multicolumn{1}{c|}{\textbf{1.45}}  & \textbf{2.24} & 1.73 & 1.66 \\
\bottomrule
\end{tabular}}
\label{tab:cross_dataset_evaluation_minFDE}
\end{table*}

\begin{table*}
\caption{\textbf{Cross-dataset generalization and multi-dataset training experiments minADE metric.}
}
\resizebox{1.0\textwidth}{!}{
\begin{tabular}{@{}l r c *{8}{@{\hspace{0.05cm}}c} @{}}
\toprule
& & \multicolumn{3}{c}{\text{MTR \cite{shi2022motion_mtr}}} & \multicolumn{3}{c}{\text{Wayformer * \cite{nayakanti2023wayformer}}} & \multicolumn{3}{c}{\text{AutoBot \cite{girgis2022autobot}}} \\
\cmidrule(lr){3-5} \cmidrule(lr){6-8} \cmidrule(lr){9-11}
& & \multicolumn{9}{c}{\cellcolor{orange!25} $\leftarrow$ \textbf{Evaluation} $\rightarrow$}\\
\cellcolor{blue!15}{$\downarrow$ \textbf{Training}} & \#\text{trajs} & \cellcolor{orange!25} \text{nuScenes}  & \cellcolor{orange!25} \text{Argoverse~2} & \cellcolor{orange!25} \text{\waymo{}} & \cellcolor{orange!25} \text{nuScenes}  & \cellcolor{orange!25} \text{Argoverse~2} & \cellcolor{orange!25} \text{\waymo{}} & \cellcolor{orange!25} \text{nuScenes}  & \cellcolor{orange!25} \text{Argoverse~2} & \cellcolor{orange!25} \text{\waymo{}}\hspace{-0.3cm}\\
\midrule
\cellcolor{blue!15} \text{nuScenes} & 32\text{k} & \textbf{1.06} & 1.85 & \multicolumn{1}{c|}{2.85} & \textbf{1.04} & 1.85 & \multicolumn{1}{c|}{2.58}& \textbf{1.21} & 1.62 & 2.35 \\
\cellcolor{blue!15} \text{Argoverse~2} & 180\text{k} & 1.42 & \textbf{0.85} & \multicolumn{1}{c|}{1.73} & 1.44 & \textbf{0.85} & \multicolumn{1}{c|}{1.65} & 1.59 & \textbf{0.85} & 1.43 \\
\cellcolor{blue!15} \text{\waymo{}  } & 1800\text{k} & 1.17 & 1.50 & \multicolumn{1}{c|}{ \textbf{0.78}} & 1.17 & 1.48 & \multicolumn{1}{c|}{\textbf{0.65}}  & 1.42 & 1.15 & \textbf{0.73} \\
\midrule
\cellcolor{blue!15} \text{All} &  2012\text{k} & \textbf{0.85} & \textbf{0.82} & \multicolumn{1}{c|}{\textbf{0.78}}  & \textbf{0.84} & \textbf{0.76} & \multicolumn{1}{c|}{\textbf{0.65}}  & \textbf{1.12} & 0.86 & 0.74 \\
\bottomrule
\end{tabular}}
\label{tab:cross_dataset_evaluation_minADE}
\end{table*}

\begin{table*}
\caption{\textbf{Cross-dataset generalization and multi-dataset training experiments miss rate metric.}
}
\resizebox{1.0\textwidth}{!}{
\begin{tabular}{@{}l r c *{8}{@{\hspace{0.05cm}}c} @{}}
\toprule
& & \multicolumn{3}{c}{\text{MTR \cite{shi2022motion_mtr}}} & \multicolumn{3}{c}{\text{Wayformer * \cite{nayakanti2023wayformer}}} & \multicolumn{3}{c}{\text{AutoBot \cite{girgis2022autobot}}} \\
\cmidrule(lr){3-5} \cmidrule(lr){6-8} \cmidrule(lr){9-11}
& & \multicolumn{9}{c}{\cellcolor{orange!25} $\leftarrow$ \textbf{Evaluation} $\rightarrow$}\\
\cellcolor{blue!15}{$\downarrow$ \textbf{Training}} & \#\text{trajs} & \cellcolor{orange!25} \text{nuScenes}  & \cellcolor{orange!25} \text{Argoverse~2} & \cellcolor{orange!25} \text{\waymo{}} & \cellcolor{orange!25} \text{nuScenes}  & \cellcolor{orange!25} \text{Argoverse~2} & \cellcolor{orange!25} \text{\waymo{}} & \cellcolor{orange!25} \text{nuScenes}  & \cellcolor{orange!25} \text{Argoverse~2} & \cellcolor{orange!25} \text{\waymo{}}\hspace{-0.3cm}\\
\midrule
\cellcolor{blue!15} \text{nuScenes} & 32\text{k} & \textbf{0.41} & 0.58 & \multicolumn{1}{c|}{0.71} & \textbf{0.42} & 0.61 & \multicolumn{1}{c|}{0.73} & \textbf{0.40} & 0.52 & 0.65 \\
\cellcolor{blue!15} \text{Argoverse~2} & 180\text{k} & 0.47 & \textbf{0.30} & \multicolumn{1}{c|}{0.59} & 0.48 & \textbf{0.28} & \multicolumn{1}{c|}{0.61} & 0.49 & \textbf{0.27} & 0.55 \\
\cellcolor{blue!15} \text{\waymo{}  } & 1800\text{k} & 0.43 & 0.44 & \multicolumn{1}{c|}{\textbf{0.22}} & 0.44 & 0.45 & \multicolumn{1}{c|}{\textbf{0.25}} & 0.42 & 0.40 & \textbf{0.25} \\
\midrule
\cellcolor{blue!15} \text{All} &  2012\text{k} & \textbf{0.32} & \textbf{0.28} & \multicolumn{1}{c|}{\textbf{0.33}} & \textbf{0.27} & \textbf{0.23} & \multicolumn{1}{c|}{\textbf{0.22}} & \textbf{0.36} & \textbf{0.27} & \textbf{0.25} \\
\bottomrule
\end{tabular}}
\label{tab:cross_dataset_evaluation_MR}
\end{table*}

\subsection{Experiments using pedestrian trajectories}
While our study specifically focuses on vehicle trajectory prediction, the UniTraj framework still supports all types of traffic participants, including cyclists and pedestrians.
To demonstrate this, we train Autobot to predict \emph{pedestrian} trajectories in \Cref{tab:pedestrian}.
Note that nuScenes is not considered as it does not officially support pedestrian trajectory prediction. 
Overall, similar findings can be made as with vehicles (Tab.\,2). 

\begin{table}[h]
\caption{\textbf{Cross-dataset generalization and multi-dataset training using pedestrian trajectories}}
\resizebox{0.8\columnwidth}{!}{
\begin{tabular}{@{}l cccccccc@{}}
\toprule
\textbf{AutoBot} & \multicolumn{2}{c}{brier-minFDE} & \multicolumn{2}{c}{minFDE} & \multicolumn{2}{c}{minADE} & \multicolumn{2}{c}{Miss Rate} \\
\colorbox{blue!15}{Train $\downarrow$} / \colorbox{orange!15}{Val $\rightarrow$} & \cellcolor{orange!15} AV2            & \cellcolor{orange!15} \waymo{}           & \cellcolor{orange!15} AV2         & \cellcolor{orange!15} \waymo{}        & \cellcolor{orange!15} AV2         & \cellcolor{orange!15} \waymo{}        &\cellcolor{orange!15}  AV2           & \cellcolor{orange!15} \waymo{}         \\ \midrule
\colorbox{blue!15}{AV2\hspace{0.5cm}}      & 1.95           & 2.40            & 1.38        & 1.75         & 0.67        & 0.85         & 0.13          & 0.16          \\
\colorbox{blue!15}{\waymo{}}    & 2.22           & 1.92            & 1.51        & 1.28         & 0.75        & 0.57         & 0.16          & 0.14          \\
\colorbox{blue!15}{All\hspace{0.7cm}}      & 1.82           & 1.92            & 1.24        & 1.28         & 0.64        & 0.57         & 0.11          & 0.14          \\ \bottomrule
\end{tabular}}
\label{tab:pedestrian}
\end{table}

\section{More details about UniTraj framework}
\label{sec:appendix_additional_details}
\subsubsection{Training pipeline}
The training pipeline within UniTraj is underpinned by PyTorch Lightning~\cite{Falcon_PyTorch_Lightning_2019}. PyTorch Lightning is a sophisticated deep-learning framework that caters to the needs of AI researchers and machine-learning practitioners. In our implementation, we utilize PyTorch Lightning's training module in Distributed Data-Parallel (DDP) mode, enabling efficient multi-GPU acceleration for enhanced training speed and effectiveness.

\subsubsection{Training settings}
In our experiments, we use 8 $\times$ A100 GPUs to train all the models. The batch sizes for MTR, Wayformer, and Autobot are 256, 256 and 128 respectively. The training takes approximately 1 hour, 15 minutes and 5 minutes per epoch for MTR, Wayformer, and Autobot, respectively. We pick the checkpoint based on the best min-brierFDE on the validation set.
